\documentclass{jmlr}

\usepackage{natbib}

\usepackage{algorithm}
\usepackage{algorithmic}

\title[A Comparison of Learning Algorithms on the Arcade Learning Environment]{A Comparison of Learning Algorithms on the Arcade Learning Environment}

  \author{\Name{Aaron Defazio} \Email{aaron.defazio@anu.edu.au}\\
  \addr NICTA/Australian National University, Canberra, ACT, Australia
 \AND
  \Name{Thore Graepel} \Email{Thore.Graepel@microsoft.com}\\
 \addr Microsoft Research Cambridge, UK
 }

\begin{document}

\maketitle

\begin{abstract}
Reinforcement learning agents have traditionally been evaluated on small toy problems. With advances in computing power and the advent of the Arcade Learning Environment, it is now possible to evaluate algorithms on diverse and difficult problems within a consistent framework. We discuss some challenges posed by the arcade learning environment which do not manifest in simpler environments. We then provide a comparison of model-free, linear learning algorithms on this challenging problem set.
\end{abstract}


\section*{Introduction}
\label{sec:introduction}

Reinforcement learning in general environments is one of the core problems in AI. The goal of reinforcement learning is to develop agents that learn, interact and adapt in complex environments, based off of feedback in the form of rewards. 

What is a  `general' environment? Agents that use raw bit streams exist at one end of the spectrum \citep{mccallum1996reinforcement}, whereas methods highly tuned for specific problem classes exist at the other. In this work we will be considering the Arcade Learning Environment (ALE, \citealt{Bell13}), which consists of a large and diverse set of arcade games. These games possess substantial structure; their state is presented to the player as low-resolution images, consisting of basic bitmap graphics.  However they are varied enough that it is difficult to program an agent that work well across games.

The diversity in the arcade learning environment poses challenges that are not well addressed in the reinforcement learning literature. Methods that work for well-worn test problems (Mountain Car, Grid-World, etc. ) can not necessarily handle the additional complexity in the Arcade Learning Environment. 

In this work we single out a number of classic and modern reinforcement learning algorithms. These methods were chosen as representative methods in their respective classes. Each of the methods we consider learns a linear policy or value function, and so they are on equal footing with regards to the representational power they possess. The restriction to linear methods is party computational and partly pragmatic. Each of the methods we consider can in principle use non-linear function approximation instead.

We also evaluated a number of high-level variations on a 5-game subset using SARSA. These include varying discounting, decay, exploration frequency and length. The results we present are a useful guidance to other researchers working on the ALE environment.

\begin{figure*}[t]
\begin{center}
\centerline{\includegraphics[width=\textwidth]{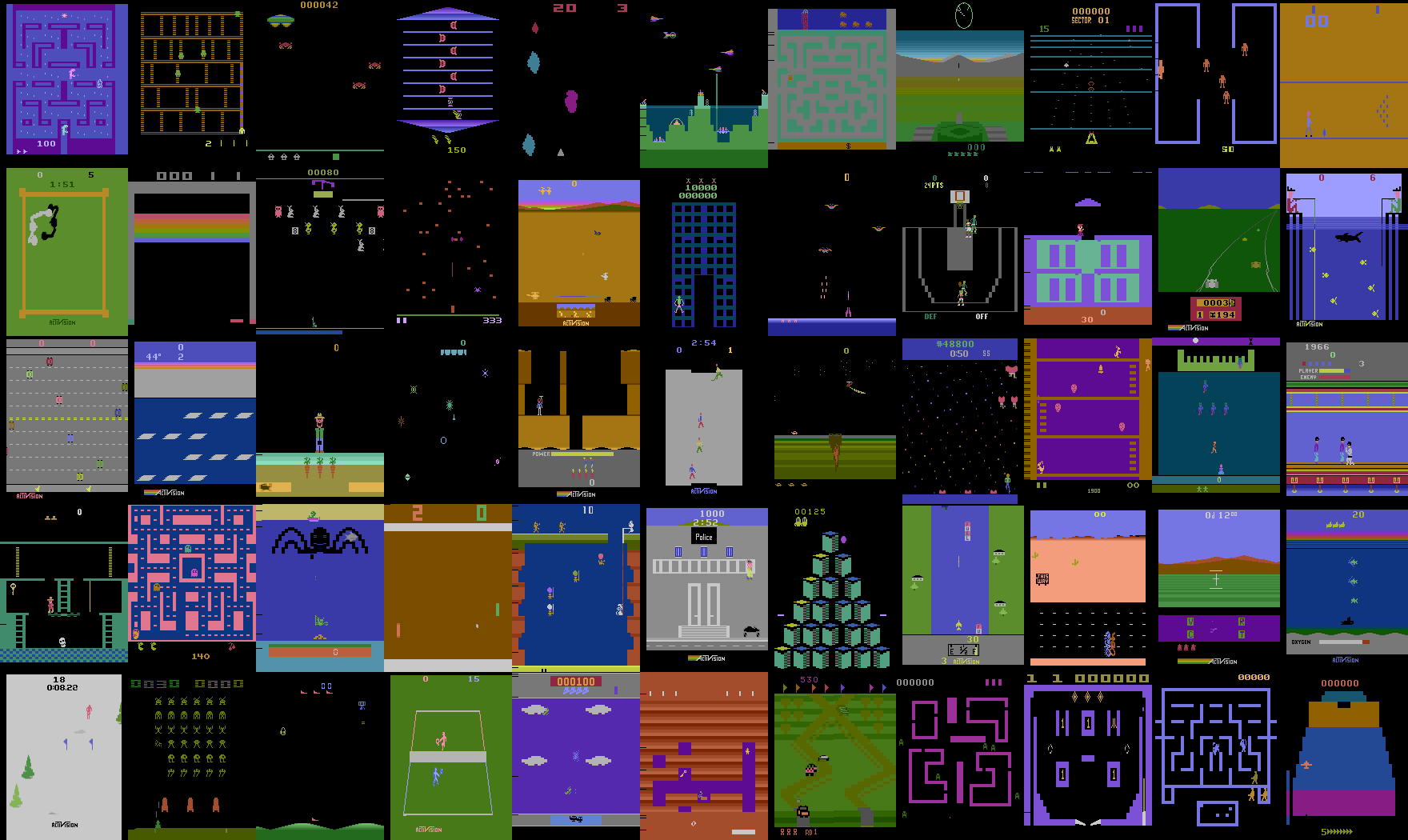}}
\caption{Screen shots from the 55 games considered in this work}
\label{fig:games}
\end{center}
\end{figure*} 

\section{Environment Overview}
\label{sec:overview}
The arcade learning environment is a wrapper around the \htmladdnormallinkfoot{Stella emulator}{http://stella.sourceforge.net/} for the Atari 2600. It wraps the emulator in a traditional reinforcement learning interface, where a agent acts in an environment in response to observed states and rewards. Atari 2600 games are remarkably well suited to being used as reinforcement learning evaluation problems; the majority of games consist of short, episodic game-play and present a score to the player which can be interpreted as a reward without issue. More modern video game consoles have longer game-play episodes, and have moved away from score based game-play.

Agents interact with ALE through a set of 18 actions, for which the actions 0-5 can be considered \emph{basic} actions, with the remaining actions being combinations of the basic actions. The basic action set consists of movement in 4 directions, firing or a NO-OP action. The state presented to the agent consists of a 210 pixel high by 160 wide screen, with each pixel a 0-127 (7 bit) color value, with the exception of a small number of games which have a slightly different screen height. The Atari 2600 console supported a reduced color space known as SECAM, which was used on early European TV sets. Mapping the 128 colors to the 8 SECAM colors gives a reduced state space size without much loss of information. 

The ALE environment supports the agent acting at a rate of up to 60 steps per second. This rate is unnecessarily rapid, and results in inhumanly jittery play for most games. We recommend following \citet{Bell13}'s suggestion and acting at 12 steps per second. This is specified by setting \texttt{frame\_skip=5} within the \texttt{stellarc} configuration file.

ALE is straight forward to build and run on Linux and Mac OS X environments. Some work is required to build on Windows, so we have published a precompiled executable on the first author's website. The ALE software supports 61 Atari 2600 games as of version 0.4. Each game has to be specifically targeted, as the game's score needs to be extracted from the game's RAM, and episode termination conditions must be identified. Of the supported games, 5 have been singled out by \citet{Bell13} as a training set for setting hyper-parameters, and 50 others as the test set. Of these games, three have non-standard screen height (Carnival, Journey Escape and Pooyan).

\section{Challenges}
\label{sec:challenges}
The arcade learning environment is substantially more complex than text-book reinforcement learning problems. It exemplifies many of the attributes found in other hard reinforcement learning problems. In this section we discuss a number of practical issues which need to be addressed by agents.

\subsection{Exploration}
\label{subsec:exploration}
The need for exploration is a well known issue that must be addressed by any reinforcement learning agent. We found that this problem is particularly exacerbated in the arcade learning environment. When using model-free methods, the following approaches are typically used:

\paragraph{Epsilon-greedy policy}
The epsilon-greedy approach, where with a small probability ($\epsilon \approx 0.05$) a random action is taken. This method has substantial trouble with the rapid acting rate in the arcade learning environment. Each single random action only slightly perturbs the state of the system, resulting in very little exploration. For example, in the game Space Invaders, the player's ship is restricted to 1 dimension of movement. Even then, the exploratory actions exhibit slow movement due to random-walk like behavior. Very rarely does the ship move from one end of the position range to the other.

More complex exploratory action sequences necessary for high-level game-play also do not occur. For example, in the game Seaquest, the player's submarine must surface for air occasionally in order to obtain human-level performance. The sequence of actions required to surface, then re-submerge, is not performed during exploration with an epsilon-greedy approach.

In order to achieve the best results in practice with on-policy RL methods, the $\epsilon$ parameter needs to be reduced over time. However, this is rarely done in published research, as determining a reduction schedule requires a tedious parameter search, and is extremely problem dependent.

\begin{figure}[t]
\begin{center}
\centerline{\includegraphics[scale=1.4]{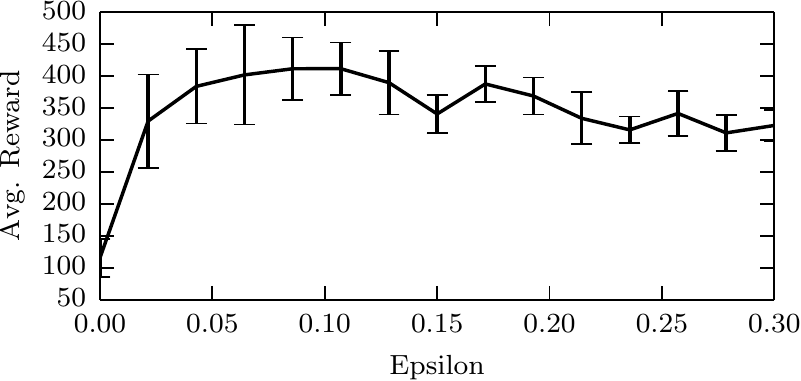}}
\caption{Effect of varying epsilon on Seaquest with a SARSA agent}
\label{fig:varying-epsilon}
\end{center}
\end{figure}

Figure \ref{fig:varying-epsilon} shows the effect of varying epsilon on Seaquest. Interesting, the standard exploration amount used in the literature ($\epsilon=0.05$) also gives good results here.  Degenerate levels ($< 0.02$) of course give poor results, but a reasonable policy is still learned for very random policies ($\epsilon>0.2$). We believe this is caused by the short time scales the agent is acting on. The noise is averaged out over the longer time-scales that game-play occurs at.

\paragraph{Softmax policy}
Another simple approach to exploration, where Q-values are used to form a Gibbs distribution over actions at each step. The chosen action is then a sample from this distribution. This approach avoids the need to choose an exploration reduction schedule, instead just requiring a scalar temperature parameter to be set. In the arcade learning environment, we found this approach to be unworkable in practice. The main issue being an extreme sensitivity to the temperature. In order for it to work at all, the temperature needs to be fine tuned for each game and each agent.

\begin{figure}[t]
\begin{center}
\centerline{\includegraphics[scale=1.4]{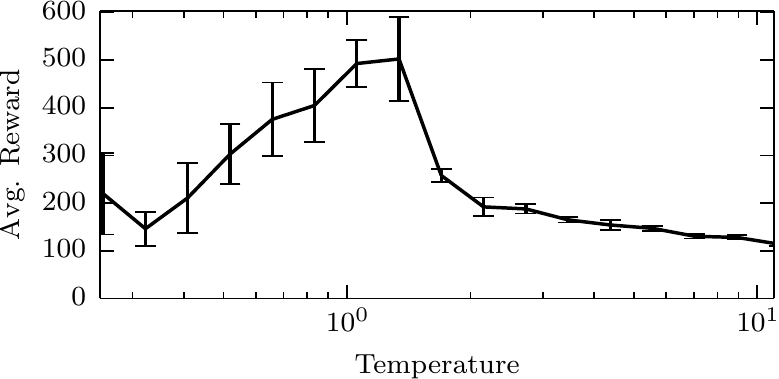}}
\caption{Effect of varying temperature on Seaquest}
\label{fig:varying-temp}
\end{center}
\end{figure}

Figure \ref{fig:varying-temp} shows the effect of varying the temperature on Seaquest. Clearly the agent is only successful for a small range of temperatures, roughly 0.8-2. None of the values we tried gave results comparable to the $\epsilon$-greedy policy at the best $\epsilon$.

\paragraph{Optimistic initialization}
Perhaps the most effective approach for simplistic grid-world like-environments is to initialize the agent so that it believes unvisited states are unreasonably good. This encourages exploration of all states at least once. In environments with a small number of states, such as grid-world mazes and the like, this is extremely effective. For the arcade learning environment, it is not clear how similar results can be achieved. The position of the agent alone can be captured fairly directly with some state encodings, but position is not the only property that needs to be explored. The matter is complicated further when non-linear value function approximation is used, as the value of states will be pushed down from their original optimistic values during learning, even before they have been visited.

\paragraph{Imitation learning}
Another approach to exploration is to guide the agent in someway using an outside `teacher'. This has been explored in depth in robotics, where it is known as imitation learning \citep{he2012imitation, price2003accelerating} or demonstration learning \citep{argall2009survey}. The simplest variant is where the reinforcement learning agent can query the teacher, to find out what action the teacher suggests it should take in any state.  A number of practical issues arise in the ALE setting however. A human teacher can not be used due to the large number of games and rate of play of the agents. It would require impractical amounts of time. Using a computerized agent observed by the learner is the other possibility. This would require domain specific AI to be programmed for each game.

\subsection{Computational Resources}
\label{subsec:comp-resources}
Stella is capable of simulating the Atari 2600 at around 7,200 frames per second on our 3.6 GHz Xeon machine; much faster than real time (60 fps). When running with the additional overhead of ALE, and communicating via pipes with a separate Python process, we found that a basic agent could act at roughly 1000 actions per second. When using \texttt{frame\_skip=5} as discussed Section \ref{sec:overview}, which is used in all the experiments detailed in this work, this is reduced to 200 actions per second. In practice, when including the overhead of the agents calculations, we were able to run our agents at approximately 130 actions per second.

Given this simulation rate, training for 5000 episodes for a single agent/environment pair usually took 1-3 days. We believe a carefully coded C++ implementation would be several times faster than this, but even then, simulations are quite computationally prohibitive. We would not recommend experimenting with ALE without access to a computing cluster to run the experiments on.

The 5,000 episode limit used by \citet{Bell13} is large but necessary. While some methods we experimented with learned reasonable parameters in a few as 500 episodes, some were still improving at the 5,000 episode limit. This is of course very much larger than what a human player requires, but without providing prior knowledge about general game mechanics and spatio-temporal relationships, a reinforcement learning agent will always be slower to learn.

\subsection{State space size}
\label{subsec:state-space-size}
Given the large size of the state space, running non-linear value function learning algorithms is also quite computationally expensive. The 210x160 screen images are similar in size to datasets used in current deep learning research (256x256), and the several million frames encountered during training is also comparable.  Even when using reduced state spaces, such as the BASIC representation detailed in \citet{Bell13}, simple neural network models are quite slow, only simulating at 1-3x real-time. 

Tile coding is the most practical feature extraction technique. We also experimented with convolutional features, where a set of predefined filters were run over the image each step. The large number of convolutions required was too slow, at least using \htmladdnormallinkfoot{OpenCV}
	{http://opencv.org/} or \htmladdnormallinkfoot{Theano}{http://deeplearning.net/software/theano/} convolutional codes.

We performed our experiments using a variant of the BASIC representation, limited to the SECAM color set. This representation is simply an encoding of the screen with a courser grid, with a resolution of 14x16. Colors that occur in each 15x10 block are encoded using indicator features, 1 for each of the 8 SECAM colors. Background subtraction is used before encoding, as detailed in \citet{Bell13}.

\section{Learning Algorithms}
\label{sec:agents}
Here we briefly outline the algorithms that we tested. We are considering the standard reinforcement learning framework; at each time step $t$ the agent observes a state $s^{(t)}$ and reward $r^{(t)}$. The action selected at each time step is denoted $a^{(t)}$. The state-action pair is processed using a state encoder function into a vector $\phi(s^{(t)},a^{(t)})=\phi^{(t)}$.   This binary vector contains the BASIC SECAM representation discussed above, of length $n$, followed by the (flattened) Cartesian product of the representation and a 1-of-18 action indicator vector. An additional bias feature which is always active was used.

We will consider algorithms that learn a linear value function associating a real number with each state-action pair ($Q_{t}(s,a)=\langle \theta^{(t)}, \phi(s,a) \rangle$), whose interpretation depends on the agent. The parameter vector $\theta$ and the corresponding function $Q$ change over time, so they are indexed by the time-step also. The parameter vector has an associated step size parameter $\alpha$. Reward discounting was used, denoted $\gamma$, with $\gamma=0.993$ used unless noted otherwise.

All agents we consider will use eligibility traces $e^{(t)}$, so that they are able to learn from delayed rewards \citep{replacingTraces}. This introduces a parameter $\lambda$, which we will discuss further in Section \ref{subsec:decay}. We implemented them in the replacing fashion, with the following update being used:
\begin{equation}
e^{(t)}_{i} = 
	\begin{cases} 
		 1 & \textrm{if } \phi^{(t)}_{i} = 1\\
		 \gamma \lambda e^{(t-1)}_{i} &\textrm{if }  \phi^{(t)}_{i} = 0\\
	\end{cases}.
\end{equation}

\paragraph{SARSA($\lambda$)}
The first algorithm we tested was SARSA \citep{SARSA}. This is perhaps the most widely used reinforcement learning algorithm. Although since its discovery other algorithms have been developed that have better theoretical guarantees, or better performance on specific problems, SARSA still gives state-of-the-art performance. SARSA is defined by the update equations:
\begin{equation}
\theta^{(t+1)} = \theta^{(t)} + \alpha \delta^{(t)} e^{(t)}
\end{equation}
where:
\begin{equation}
 \delta^{(t)}= r^{(t+1)} + \gamma  Q_{t}(s^{(t+1)},a^{(t+1)}) -  Q_{t}(s^{(t)}, a^{(t)}).
\end{equation}

\paragraph{Q($\lambda$)}
We also tested against Q-learning \citep{QlearningWatkins1989}, the off-policy alternative to SARSA. It differs by the update for $\delta$:
\begin{equation}
\delta^{(t)}= r^{(t+1)} + \gamma  \max_{a^\prime} Q_{t}(s^{(t+1)},a^{\prime}) -  Q_{t}(s^{(t)}, a^{(t)})
\end{equation}
Q-learning can potentially learn a better policy than SARSA in games where death can easily be caused by the random moves included in the $\epsilon$-greedy policies learned by on-policy methods. The downside is a stochastic policy can give better results in some games. Off-policy methods are also known to diverge in some cases when function approximation is used \citep{baird}.

\paragraph{ETTR($\lambda$)}
We also implemented a \emph{shorted-sighted} agent, that aims to minimize the expected time to next positive reward (ETTR), instead of the discounted expected future reward optimized by the other agents. This has the advantage of potentially being easier to learn, as it gets a non-noisy signal whenever it actually reaches a positive reward. The disadvantage is a lack of long term planning and poorer risk-aversion. Within the temporal difference framework, with decaying eligibility traces, ETTR uses the following update for $\delta$:
\begin{equation}
	\delta^{(t)} = 
	\begin{cases} 
		- Q_{t}(s^{(t)},a^{(t)})  + 1   \\
			 \quad + Q_{t}(s^{(t+1)},a^{(t+1)}) & \textrm{if }r^{(t+1)} = 0 \\
		- Q_{t}(s^{(t)},a^{(t)})  & \textrm{if }r^{(t+1)} > 0 
	\end{cases}.
\end{equation}
No discounting is used. We only applied ETTR to games where the reward structure contained positive rewards as the primary motivator of the agent. This excluded 10 of the 55 games, mainly games where negative reward was given until a goal was achieved (Boxing, Double Dunk, Enduro, Fishing Derby, Ice Hockey, Journey Escape, Pong, Private Eye, Skiing, Tennis). These games could be approached by a similar scheme, where expected length of episode is minimized instead.

\paragraph{R($\lambda$)}
Another class of reinforcement learning agents seek to optimize the expected reward per time step instead. R-learning is the primary example of such a method in the off-policy case \citep{rlearning}. Such a formulation avoids the need for discounting, but introduces an additional step size parameter $\beta$, which controls the rate at which an estimate of the expected reward per time-step $\rho$ changes. The update equations are:
\begin{equation}
\delta^{(t)} = r^{(t+1)} - \rho^{(t)} 
+  \max_{a^\prime} Q_{t}(s^{(t+1)},a^{\prime}) -  Q_{t}(s^{(t)},a^{(t)}) ,
\end{equation}
where $\rho$ is updated whenever we take on policy actions, with:
\begin{equation}
\begin{split}
\rho^{(t+1)} ={} & \rho^{(t)} +  \beta \left( r^{(t+1)} - \rho^{(t)} \right. \\
  + & \max_{a} Q_{t}(s^{(t+1)},a)
 - \left. \max_{a^\prime} Q_{t}(s^{(t)},a^\prime) \right).
\end{split}
\end{equation}
Expected reward methods have seen more use in the Actor-Critic setting (see below).

\paragraph{GQ($\lambda$)}
The \emph{gradient temporal difference} (GTD, \citet{gtd}) class of algorithms are an attempt to improve the convergence properties and robustness of classical temporal difference algorithms. They phrase learning as a stochastic gradient descent problem, for which stronger convergence properties are known. We applied the {GQ($\lambda$) algorithm \citep{gq} to the ALE environment. GTD-style algorithms maintain a weight vector $w$ as well as a parameter vector $\theta$. Both vectors are updated at each step, with step sizes $\alpha$ and $\beta$ respectively. The update equations are modified as:
\begin{equation}
\begin{split}
\delta^{(t)}& = r^{(t+1)} + \gamma E_{a} \left[ Q_{t}(s^{(t+1)},a) \right]  -  Q_{t}(s^{(t)}, a^{(t)}), \\
\theta^{(t+1)} & = \theta^{(t)} + \alpha \left(
\delta^{(t)} e^{(t)} - \gamma(1-\lambda) \langle w^{(t)}, e^{(t)} \rangle
 \bar{\phi}^{(t+1)} \right), \\
w^{(t+1)} & = w^{(t)} + \beta \left(  \delta^{(t)} e^{(t)} -  
	\langle w^{(t)}, \phi^{(t)} \rangle \phi^{(t)}  \right).
\end{split}
\end{equation}
The expectation over actions is taken with respect to the epsilon-greedy policy.


\paragraph{Actor-Critic}
Actor-Critic methods \citep{actorcritic} decouple the policy and value function predictors. The \emph{critic} is used for value function estimation, and the \emph{actor} maintains the policy. We will consider the simplest case, where both are linear. The key advantage of (linear) actor critic methods is that we can use different learning rate parameters for the actor and the critic. Typically the actor (step length $\beta$) is set to evolve slower than the critic (step length $\alpha$), so that the value function estimates have more time to stabilize as the policy changes. We tested a non-standard variant that uses epsilon-greedy policies, rather than the Gibbs/Boltzmann methods more common in the literature. This choice was made based on the results in Section \ref{subsec:exploration}.
Let $\nu$ denote the critic's weight vector. Then $\delta$ remains the same as for SARSA, but with the value function approximation changed to  ($Q_{t}(s,a)=\langle \nu^{(t)}, \phi(s,a) \rangle$). The parameter updates are:
\begin{equation}
\begin{split}
\nu^{(t+1)} &= \nu^{(t)} + \alpha \delta^{(t)} e^{(t)}, \\
\theta^{(t+1)} &= \theta^{(t)} + \beta \delta^{(t)} e^{(t)}. \\
\end{split}
\end{equation}
Action selection by the actor is done as with SARSA, by taking the action that maximizes $\langle \theta^{(t)}, \phi(s^{(t)},a) \rangle$.

\paragraph{Other methods}
Several families of methods could not be tested due to computational considerations. We briefly discuss some of them here. Least-squares methods such as LSPI \citep{LSPI} are perhaps the most prominent alternative within the class of linear value function approximation methods. Unfortunately, they require quadratic time operations in the state vector size. Even though we are using the smallest screen based state representation considered in the literature for ALE, the quadratic time operations were too slow for us to use.

There is also a large literature on methods that store past visited state/action information, the main example being Fitted Q-iteration \citep{fittedq}. Given the millions of states visited during training, only a small subset of the history can tractably be stored for ALE. Investigating tractable variants of history-based methods for ALE is an interesting avenue of research, but outside the scope of this work.

\section{Results}
\label{sec:results}
The results in this Section were computed by averaging a number of trials. Each trial consisted of running 5000 episodes while in training mode, then 500 episodes in test mode where the learned policy is executed. For on-policy methods, the test episodes included the exploration steps, and for off-policy, the test policy was purely greedy.

\subsection{Discounting Comparison}
Figure \ref{fig:discounting-comp} shows the effect of differing discounting amount on the training set games for SARSA. The optimal discounting factor is not consistent between games, with the best value differing quite substantially between Seaquest and Asterix for example. A general trend is apparent; using extremely high discounting ($0.5$-$0.9$) is ineffective, and too little discounting ($> 0.9999$) also gives poor results. We encountered convergence problems when no discounting was used, so a value of $\gamma=1$ is omitted from the plot. The roughly convex dependence on discounting is consistent with the literature on simpler environments.

\begin{figure}[t]
\begin{center}
\centerline{\includegraphics[scale=1.4]{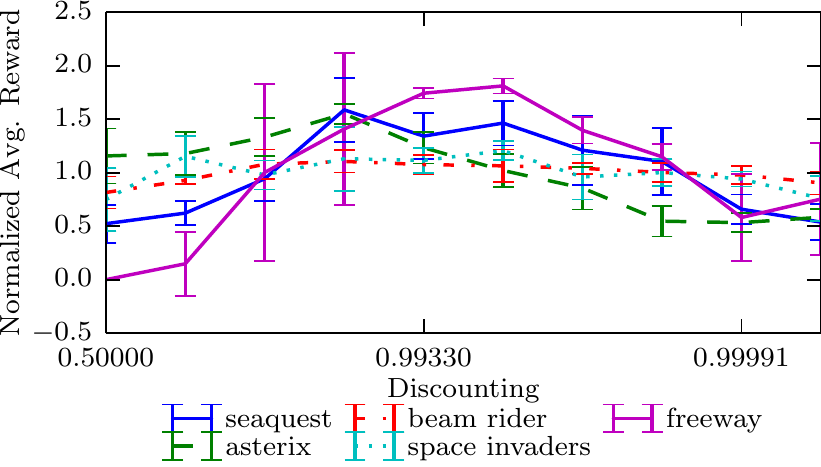}}
\caption{Effect of varying discounting on training set games}
\label{fig:discounting-comp}
\end{center}
\end{figure} 

\subsection{Decay Comparison}
\label{subsec:decay}
Figure \ref{fig:varying-decay} shows the effect of varying decay on the training set games for SARSA. There is no clear best decay, although values of both 0 and 1 are poor. Decay in the range 0.5-0.9 seems most effective. There is less sensitivity to the exact value of the decay parameter $\lambda$ than the discounting parameter $\gamma$.

\begin{figure}[t]
\begin{center}
\centerline{\includegraphics[scale=1.4]{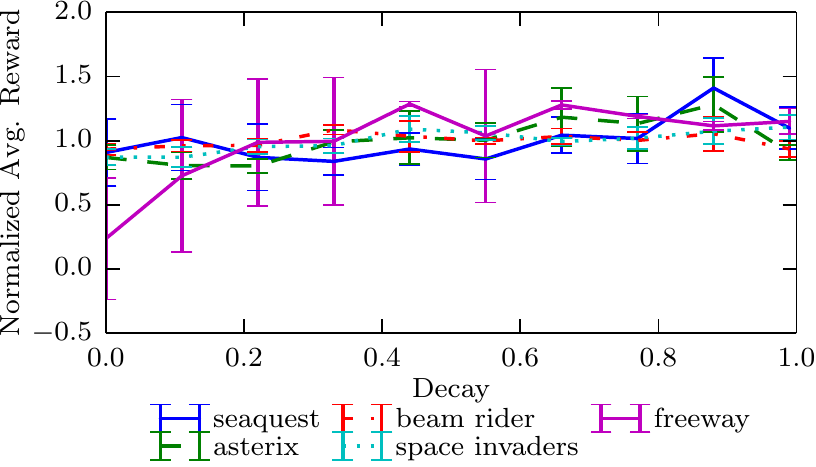}}
\caption{Effect of varying decay on training set games}
\label{fig:varying-decay}
\end{center}
\end{figure} 

\subsection{Exploration Periods, on/off policy}
We experimented with policies that extend the amount of exploration on Seaquest. We tested a modified $\epsilon$-greedy approach, where whenever the random policy was chosen, that action was then taken for more than a single step. We varied this exploration length between 1 and 6 steps. We got somewhat mixed results. Figure \ref{fig:el1} shows that for SARSA and Q-learning, depending on the value of $\epsilon$, length two exploration periods can be better than standard $\epsilon$-greedy, but the difference is close to the noise floor. Otherwise there is a downward trend in the average reward obtained, with a small up-tick around length six.

\begin{figure}[t]
\begin{center}
	\includegraphics[scale=1.4]{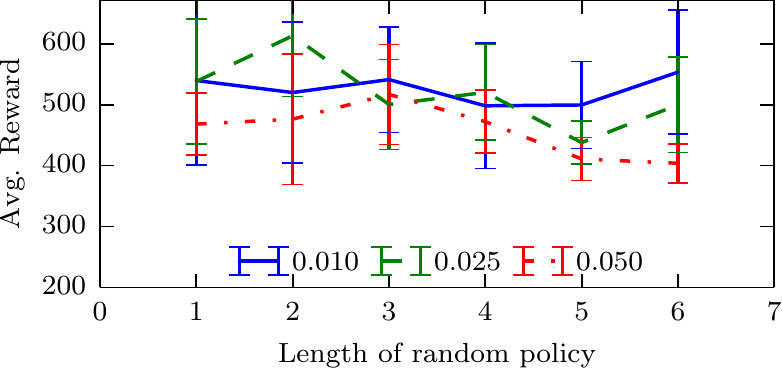}
	\includegraphics[scale=1.4]{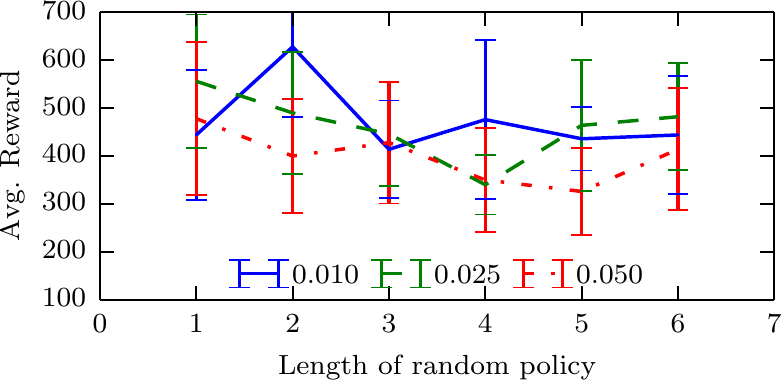}
\caption{Longer exploration periods with SARSA on Seaquest for three $\epsilon$ values, for SARSA (top) and Q-Learning (bottom)}
\label{fig:el1}
\end{center}
\end{figure} 

\subsection{Algorithm Comparison}
For each algorithm described in Section \ref{sec:agents}, we ran 5 trials on each of the games where it could be applied. Two issues arose that needed to be addressed. Several game \& algorithm combinations exhibited behaviour where the agent would make no progress and consistently hit the episode limit with zero reward. This happened in particular on Zaxxon and Montezuma's revenge. These cases were excluded from the comparisons. We also encountered divergent behaviour on Q-learning roughly 40\% of the time, and on GQ a surprising 70\% of the time. Runs where divergence occured were excluded. 

We consider SARSA as the baseline. First we compare the performance and consistency of the methods. For each algorithm we computed the average reward during the test episodes for each trail and game. We then computed the relative performance of the method as the average over each game of the ratio of the trial average against the baseline. The average was computed over the middle 90\% of games to remove outliers. We also computed the test reward standard deviation (SD) across trails for each game as a measure of consistency. The median (Q2) and 1st and 3rd quartiles of the SD across games is given together with the relative performance in Table \ref{tbl:perf}.

\begin{table}[h!]
\center
\begin{tabular}{ l | c | c | c | c}
Algorithm & Rel. perf & SD Q1 & SD Q2 & SD Q3 \\
SARSA & 1.00 & 0.26 & 0.41 & 0.55 \\
AC & 0.99 & 0.23 & 0.45 & 0.58 \\
ETTR & 1.03 & 0.32 & 0.44 & 0.58 \\
GQ & 0.65 & 0.14 & 0.42 & 0.81 \\
Q & 0.82 & 0.27 & 0.50 & 0.85 \\
R & 0.96 & 0.26 & 0.42 & 0.59
\end{tabular}
\caption{Algorithm performance and consistency comparison}
\label{tbl:perf}
\end{table}

We can see that SARSA, R, AC and ETTR all performed similarly in the relative performance metric. This suggests that the additional complexity of AC and the narrower focus of ETTR did not help. Our hypothesis that ETTR would perform more consistently is also not supported by the standard deviation results. The off-policy methods Q-learning and GQ-learning did not perform well here, showing significantly worse results.

The large variances we see can potentially effect the relative performance figures significantly. A more robust method of comparing a pair of algorithms is to count the number of games for which the average test reward is higher for one game than the other. Table \ref{tbl:pairwise} contains these comparisons. Cases where one or both games failed to converge or timed-out are excluded from the table.

\begin{table}[h!]
\center
\begin{tabular}{ l | c | c | c | c | c | c}
&  SARSA & AC & ETTR & GQ & Q & R \\
SARSA & -  &  28/25  &  22/21  &  49/5  &  44/10  &  30/23 \\
AC & 25/28  &  -  &  18/25  &  49/4  &  38/17  &  26/25 \\
ETTR & 21/22  &  25/18  &  -  &  40/1  &  36/8  &  22/20 \\
GQ & 5/49  &  4/49  &  1/40  &  -  &  8/37  &  5/49 \\
Q & 10/44  &  17/38  &  8/36  &  37/8  &  -  &  13/41 \\
R & 23/30  &  25/26  &  20/22  &  49/5  &  41/13  &  -
\end{tabular}
\caption{Pairwise higher/lower test reward comparison}
\label{tbl:pairwise}
\end{table}

The results in this table support the inferences made using the relative performance measure. SARSA appears better than R-learning by a larger margin, possibly due to chance, but otherwise the on-policy methods show similar results.

We also computed the correlation coefficient between each pair of algorithms, considering only trials for which both methods finished. The results are shown in Table \ref{tbl:corr}.
\begin{table}[h!]
\center
\begin{tabular}{ l | c  | c | c | c | c | c}
&  SARSA & AC & ETTR & GQ & Q & R \\
SARSA & 1.00 & 0.27 & -0.12 & -0.86 & -0.46 & 0.38 \\
AC & 0.27 & 1.00 & -0.27 & -0.75 & -0.10 & -0.06 \\
ETTR & -0.12 & -0.27 & 1.00 & -0.69 & -0.76 & -0.21 \\
GQ & -0.86 & -0.75 & -0.69 & 1.00 & 0.71 & -0.79 \\
Q & -0.46 & -0.10 & -0.76 & 0.71 & 1.00 & -0.07 \\
R & 0.38 & -0.06 & -0.21 & -0.79 & -0.07 & 1.00
\end{tabular}
\caption{Algorithm correlation}
\label{tbl:corr}
\end{table}

As would be expected, there is little correlation between SARSA and ETTR, suggesting they perform well on differing problems. There is moderate correlation between SARSA and AC, which would be expected given there similarities. Q and GQ learning are highly correlated, whereas they are both negatively correlated with R-learning. This suggests the behaviour of R-learning is closer to the on-policy methods, given it's correlation with SARSA.

One of the most relevant properties of a learning algorithm is its speed of convergence. For each method, we computed the average reward over the last 500 episodes of the training episodes, and compared it to the average across the 500 episodes preceding those. If the average was within 10\%, we considered the method to have converged. The percentage of converged trails out of those that finished for each method was: SARSA: 85\%; AC: 80\%; ETTR: 84\%; GQ: 80\%; Q: 82\%; R: 85\%. The convergence rates are very similar across the methods considered. These results also suggests that the step size constants that we chose were reasonable.

\section{Discussion}
\label{sec:discussion}

\subsection{Instability}
\label{sec:instability}
We were surprised by the variability in the results we saw. The same agent with the same hyper-parameters can exhibit learning curves that show little resemblance to one another. On games with fairly stable behaviour, there is often a $\pm 50 \%$ swing in results between trials. On less stable games such as Zaxxon, the agents could get stuck on degenerate low reward policies. This makes comparisons between agents difficult. It is unsatisfying to average over such effects, as it hides the true nature of what is being learned.

\subsection{Divergence}
\label{sec:divergence}
Off-policy methods such as Q-learning are known to have convergence issues when function approximation is used, even in the linear case that we considered. This is normally considered a theoretical issue when $\epsilon$-greedy policies are used, as the difference between $\epsilon$-greedy and pure on-policy is minor. However we did see a high probability of divergence in practice with Q-learning, as detailed in the results section. Surprisingly, we also experienced divergence with GQ($\lambda$), despite theoretical convergence results. Our implementation did not use the projection step that they detail. Their theory requires the projection, however they suggest it is not needed in practice. We believe that it is the likely cause of the convergence issues, combined with perhaps a sensitivity to the step parameters.

Interestingly, The off-policy R-learning method did not have the same divergence issues, and it performed nearly as well as SARSA in our experiments. It appears the best choice among the off-policy methods considered.

\subsection{Exploration}
\label{sec:expl-discuss}
Even within the class of linear value function approximation, we clearly see in our experiments that sub-optimal policies are being learned in many trials. Plotting the reward per episode over time shows that the average reward per episode stops improving, but at a level below the optimal reward level, and at a level varying between trials. We speculate that the epsilon-greedy policy is not resulting in much exploration, rather its effect is to introduce jitter that helps prevents degenerate policies from being followed, such as repeated ineffective actions, or behavioural loops.
Based on the need for large learning rates, it seems that the noise introduced into the value function by taking large steps in parameter space is contributing far more to exploration.

\subsection{Learned policies}
\label{sec:learned-policies}
Looking at raw point scores does not give a clear indication of what a policy is actually doing. To get a better idea of the sorts of policies that were being learned we examined videos of the each agent's gameplay on a number of games. For the majority of games we examined, the learned policies consisted of a very simple sequence of movements together with holding down the fire key. For example, with Seaquest, holding fire together with alternating left-down and right-down actions appeared to be the policy learned by SARSA most of the time. Similar behaviour was learned on Krull, where repeatedly jumping gives a very high score. 

We noticed that the linear algorithms we considered were capable of learning to move to a particular location on the screen at the start of the episode, then from there repeating a set of actions. For example, in Gopher, the AC agent learned to move to a fixed point right of the starting location, then to repeatedly trigger the fire action. This appeared to be effective due to a simplistic enemy AI. For Zaxxon, you die after 10 seconds with a score of zero unless you move to the center of the screen, away from the starting position on the left. This turned out to be an extremely difficult policy to learn, as the negative reinforcement (death) occurs long after the actions. All the algorithms we looked at had some runs where they completely failed to learn that policy, or discovered it then later abandoned it part way through training.

\subsection{Choosing the best algorithm for an environment}
The comparison on diverse environments we have performed allows us to see which environments each algorithm is best at and which it is worse at. First we consider the off-policy algorithms. They tended to perform best on problems where exploration actions could have major negative effect on the outcome, which is consistent with our expectations. For example, for Zaxxon mentioned above, they were more consistently able to stay in the safe region of the screen. Looking in particular at Q-learning, we see a large improvement in performance when switching from the training regime to the test regime (where no epsilon randomness is used) for Seaquest, Assault, Time pilot, Asterix, and smaller improvements for many other games. For some games, a sharp drop in performance occurs, such as for Tutankham and Space Invaders. 

Games that require a level of random left and right movement were particularly a problem, such as the Crazy Climber game, where the player just has to move forward most of the time, but occasional left or right movement is required to go around obstacles. Potentially a soft-max policy could avoid these issues, but as mentioned in Section \ref{subsec:exploration}, the temperature parameter needs to be fine-tuned separately for each game due to the large differences in score magnitudes.

The ETTR method was able to outperform the other algorithms for a number of games, including Gopher, Q*bert and Road Runner. These games had a common pattern where you need to react to movement in squares adjacent to the player character. This pattern was shared by a number of other games also, so its not clear if any conclusions can be drawn from that. The ETTR agent had significantly less risk aversion than the other algorithms considered, which was the expected behaviour. For example on Seaquest, instead of moving the player's submarine to the bottom of the screen which is the safest location, it stayed near the top of the screen where more rewards are available. In terms of actual scores, we didn't see significant negative effects from the lack of risk aversion in the games we looked at.

The AC agent was able to outperform all the other methods on a number of problems also. It had significantly better scores on Boxing, Alien and Frostbite. The AC method can also be tuned to a greater degree than SARSA, so it is a good choice when tuning on a per-problem basis is used. Additional tuning can also be a downside, although based on it's consistent performance that doesn't appear to be the case here.

\section{Related Work on the Arcade Learning Environment}
\label{sec:related-work}
The arcade learning environment is relatively new, so little work so far directly targets it. \citep{Bell13} introduces the environment. They consider the performance of several state representations with SARSA. 

\citet{Naddaf2010} presents some tables comparing decay and learning rates for 4 of the training set games. Our plots in Section \ref{sec:results} are more comprehensive. \citet{deepmind} consider the use of deep learning methods on a subset of 7 games, showing super-human performance on 3 games.

\section*{Conclusion}
\label{sec:conclusion}
The results in this paper provide a guide other researchers working with the Arcade Learning Environment. We give guidance on most effective decay, discounting and epsilon-greedy constants. We also provide a comparison of standard RL algorithms, showing that a set of commonly used linear on-policy algorithms give similar performance. We also show that some common off-policy methods have serious issues with the complexity of the ALE environment, and we give recommendations based on our empirical results on the RL algorithms that should used in different game environments.


\bibliography{rlpaper}

\end{document}